\documentclass[10pt,twocolumn,letterpaper]{article}

\usepackage{cvpr}              

\usepackage{graphicx}
\usepackage{amsmath}
\usepackage{amssymb}
\usepackage{times}
\usepackage{epsfig}
\usepackage{enumitem}
\usepackage{lipsum}

\newcommand\blfootnote[1]{%
  \begingroup
  \renewcommand\thefootnote{}\footnote{#1}%
  \addtocounter{footnote}{-1}%
  \endgroup
}

\usepackage[pagebackref,breaklinks,colorlinks]{hyperref}

\usepackage[capitalize]{cleveref}
\crefname{section}{Sec.}{Secs.}
\Crefname{section}{Section}{Sections}
\Crefname{table}{Table}{Tables}
\crefname{table}{Tab.}{Tabs.}

\def\cvprPaperID{11} 
\def\confName{CVPR}
\def\confYear{2023}

\begin{document}

\title{SEM-POS: Grammatically and Semantically Correct Video Captioning}

\author{Asmar Nadeem\(^1\), Adrian Hilton\(^1\), Robert Dawes\(^2\), Graham Thomas\(^2\), Armin Mustafa\(^1\)}


\maketitle
\begin{abstract}
  Generating grammatically and semantically correct captions in video captioning is a challenging task. The captions generated from the existing methods are either word-by-word that do not align with grammatical structure or miss key information from the input videos. 
  To address these issues, we introduce a novel global-local fusion network, with a Global-Local Fusion Block (GLFB) that encodes and fuses features from different parts of speech (POS) components with visual-spatial features. We use novel combinations of different POS components - 'determinant + subject', 'auxiliary verb', 'verb', and 'determinant + object' for supervision of the POS blocks - Det + Subject, Aux Verb, Verb, and Det + Object respectively.
  The novel global-local fusion network together with POS blocks helps align the visual features with language description to generate grammatically and semantically correct captions. 
  Extensive qualitative and quantitative experiments on benchmark MSVD and MSRVTT datasets demonstrate that the proposed approach generates more grammatically and semantically correct captions compared to the existing methods, achieving the new state-of-the-art. Ablations on the POS blocks and the GLFB demonstrate the impact of the contributions on the proposed method.
\end{abstract}

\blfootnote{\(^1\)Centre for Vision, Speech and Signal Processing (CVSSP), University of Surrey, United Kingdom. \(^2\)BBC Research and Development,
United Kingdom. Correspondence to: Asmar Nadeem
\( \mathnormal{<} \)asmar.nadeem@surrey.ac.uk\( \mathnormal{>} \).}

\section{Introduction}
\label{sec:intro}

Video captioning is a challenging task as it aims to describe the contents of a video in a natural language like humans do \cite{venugopalan2015sequence}. This finds application in media production \cite{alayrac2016unsupervised}, visual retrieval \cite{song2018quantization ,li2017general, zhao2018cnn, hong2015learning, wang2017survey}, visual question answering \cite{antol2015vqa}, etc. Existing methods for video captioning can broadly be classified into two categories: (1) learning visual representation for the whole caption in an end-to-end manner leaving the fine-grained local features unaddressed \cite{pei2019memory, aafaq2019spatio, ryu2021semantic, pan2016jointly}, which generates captions with key information missing in the description;  and (2) incorporating features at the level of objects and motion \cite{pan2020spatio, wang2019controllable, zhang2019object, zheng2020syntax, ye2022hierarchical}, generating word-by-word descriptions which do not align with the grammatical structure.
\begin{figure}[t]
  \centering
   \includegraphics[width=\linewidth]{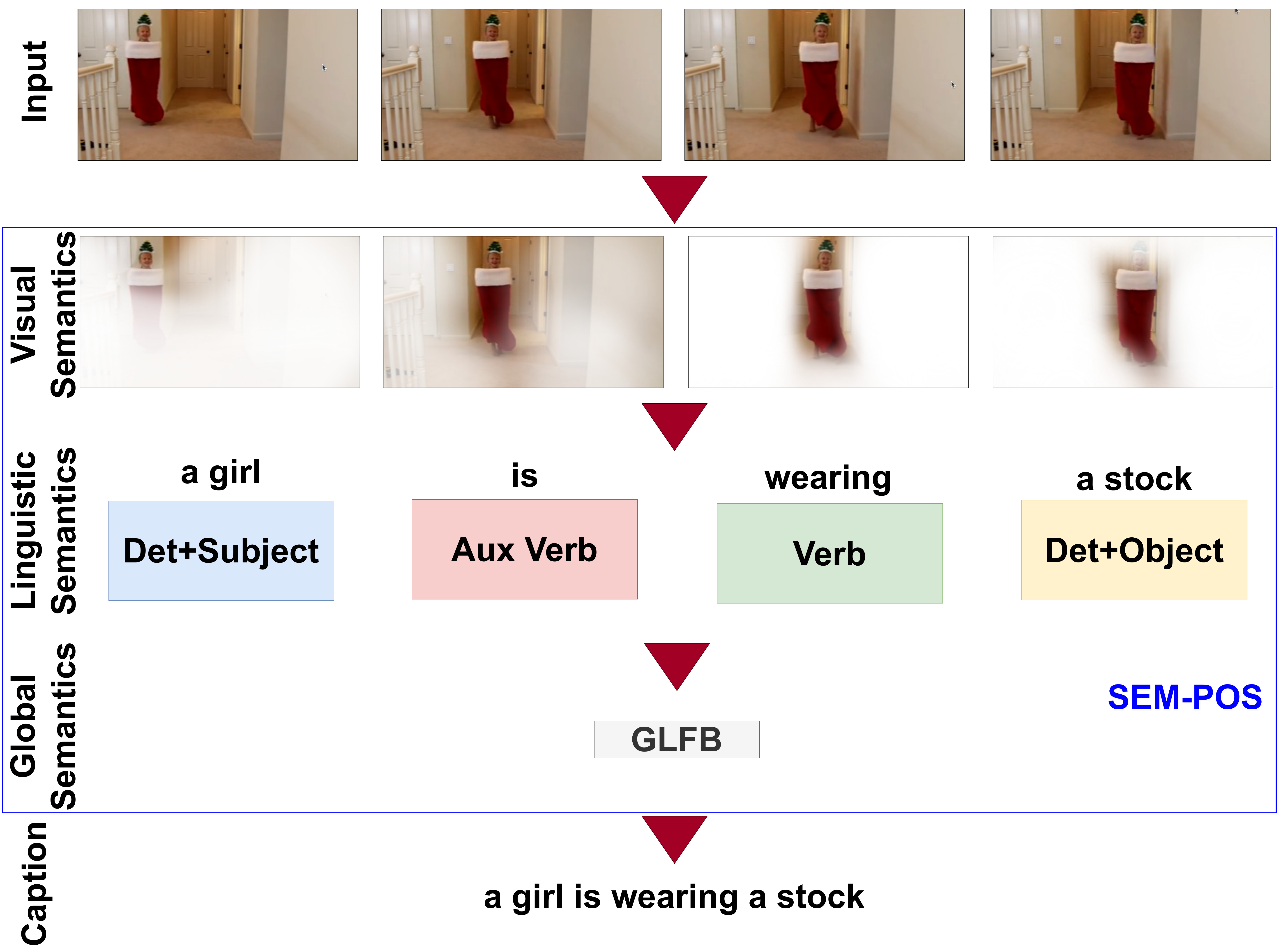}
   \caption{SEM-POS aligns visual and linguistic semantics for video captioning task. Det + Subject, Aux Verb, Verb, and Det + Object are the four POS blocks used in the proposed approach.}
   \label{fig:motivation}
\end{figure}
In this paper, we aim to address both issues in the existing methods by proposing a novel global-local fusion network that consists of- first a variety of POS components of a sentence; and second the Global-Local Fusion Block (GLFB) which fuses the local and global, visual and language features in the network. Prior knowledge of what to predict in a template-based sentence \cite{guadarrama2013youtube2text, kojima2002natural, krishnamoorthy2013generating}, gives POS blocks the ability to focus on the most suitable visual features to produce grammatically correct captions.\\
\noindent
Existing methods \cite{pei2019memory, ryu2021semantic, yan2022gl} use a single type of visual feature i.e., spatial or temporal, as an input to the entire model which leads to reduced semantic correctness.  The proposed method inputs both spatial and temporal input features for each POS block, which improves the semantics of the generated caption. The output visual features from each POS block are input to the GLFB block along with the language features, which fuses the local and global language and visual information, unlike previous methods \cite{pan2016jointly,aafaq2019spatio} that generate only a global caption from a video. This helps to align the visual semantics of a sentence with natural language semantics and grammatical structure.
\noindent
Existing methods \cite{aafaq2019spatio, pan2020spatio, zhang2019object, ye2022hierarchical} mainly use global verbs and nouns in the form of actions and objects, generating word-by-word descriptions. \cite{shen2017weakly} attempts to explore the local semantics in the input video, however it does not align the visual and language semantics, leading to limited performance. In this paper, in addition to the visual features from the POS blocks, we also input language semantics (supervised by 'determinant + subject', 'auxiliary verb', 'verb', and 'determinant + object') to the GLFB block, which aligns visual semantics with linguistics and generating grammatically correct captions. The input, output, and pipeline from the proposed network are shown in Figure \ref{fig:motivation}.  We also mask input spatial as well as temporal features in end-to-end training which  also results in an improvement in the performance. \\
\noindent
The proposed method is inspired by the previous work POS-CG \cite{wang2019controllable} and SAAT \cite{zheng2020syntax}. However, these methods do not fuse the language and visual information and they assume the subjects as well as the objects to be physical entities and the verbs to be motion-based actions which may not be true semantically. This paper aims to answer three main questions previously unaddressed by state-of-the-art methods: 1) How to align visual and linguistic semantics to generate grammatically correct captions? (2) How to add fine-grained local features to increase semantic correctness? and (3) How to fuse local and global features together for simultaneous grammatical and semantic correctness? To answer these questions, we have proposed the following novel contributions:

\begin{itemize}[topsep=0pt,partopsep=0pt,itemsep=0pt,parsep=0pt] 
	\item A novel global-local fusion network with GLFB that encodes and fuses fine-grained local visual features from different POS blocks with global caption language features for semantically correct captions.
	\item We propose four POS blocks: Det+Subject, Aux Verb, Verb and Det+Object with spatial and temporal features as input. Comprehensive ablation studies on the selection of POS blocks and input features show generation of captions closer to the natural language.
	\item An extensive evaluation of the proposed method against state-of-the-art on two benchmark datasets, MSVD \cite{chen2011collecting} and MSRVTT \cite{xu2016msr}, on five metrics, demonstrates improved grammatical and semantic caption generation. 
\end{itemize}

\begin{figure*}[t]
  \centering
   \includegraphics[width=\linewidth]{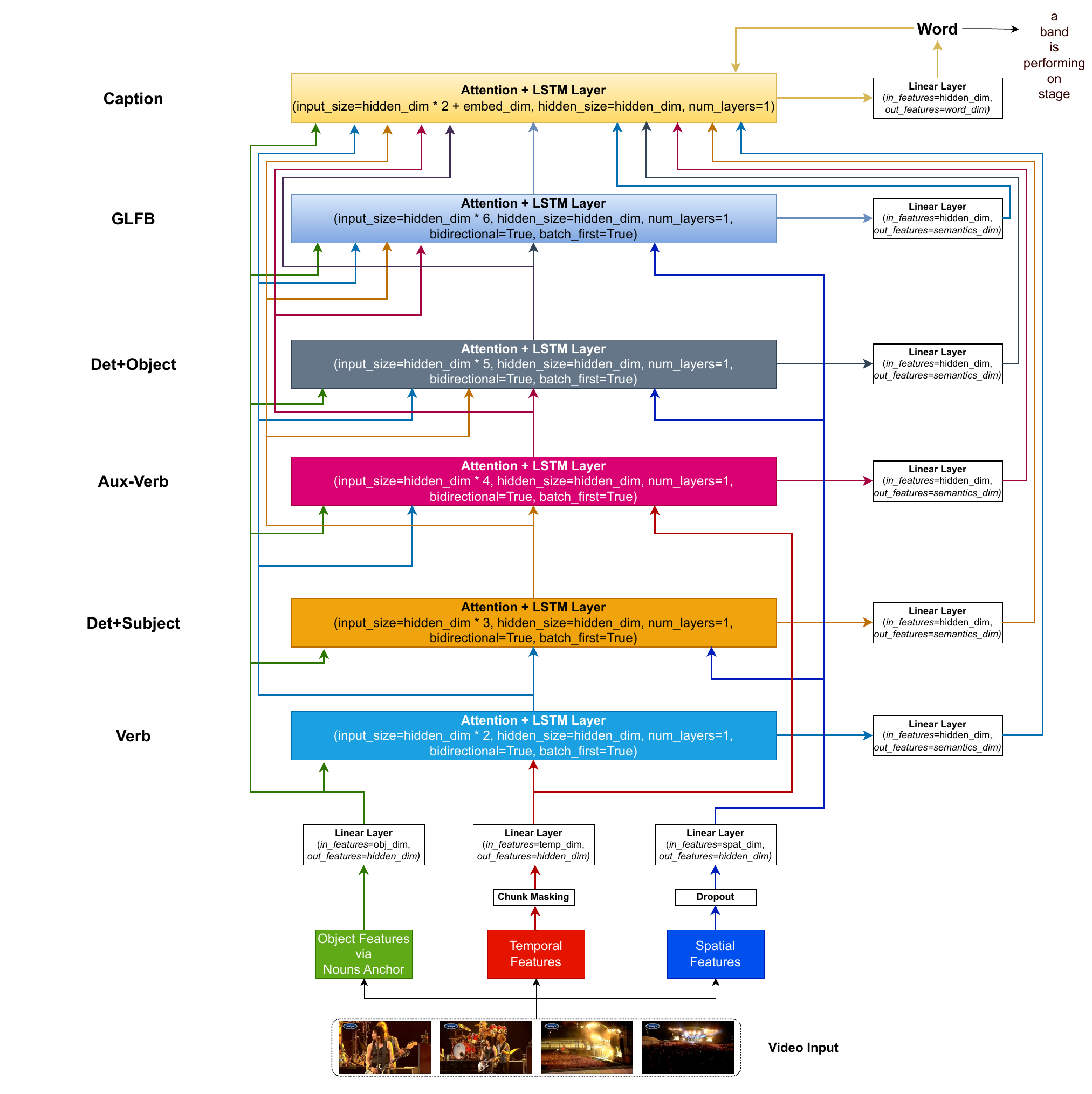}\vspace{-5mm}

   \caption{Illustration of our SEM-POS where POS blocks in Section \ref{ssec:pos_block} help align visual-linguistic semantics and attain fine-grained features. Also, GLFB (see Section \ref{ssec:glcr_block}) performs global-local feature fusion to generate grammatically and semantically correct captions.\vspace{-5mm}}  
   \label{fig:2}
\end{figure*}

\section{Related Work}
This section reviews video captioning techniques, and visual and linguistic feature fusion to support the contributions of the paper. 

\subsection{Video Captioning}
\label{sec:videorepresentation}

\noindent
\textbf{Video representation:}
Video captioning \cite{aafaq2019video, venugopalan2015sequence} is a challenging task in video representation. Earlier, CNN-based methods \cite{simonyan2014very, tran2015learning} have been applied to learn video representations which lack to model the long-range dependencies but some \cite{szegedy2017inception, hara2018can} are still popular for their spatial and temporal representations of the videos. These video features, in a global fashion, combined with RNN \cite{elman1990finding, cho2014learning, hochreiter1997long, zeng2016generation, long2018video} and LSTM \cite{gao2017video, song2017hierarchical}, have been employed for the video captioning task but visual-linguistic alignment and fine-grained features are not explored.\\
Recently, multi-modal methods \cite{xu2017learning, song2018deterministic} have become popular using text and audio information for video understanding tasks which results in text or audio-dependent models lacking visual information. Some methods \cite{hao2017end} use cross-attention mechanisms to attend to different modalities and other, vision only, methods use attention mechanisms \cite{zhao2019cam, song2017hierarchical} for extracting linguistic semantics, globally, from the visual features. 
The proposed method in this paper exploits the power of LSTMs, inspired by \cite{gao2017video}, for aligning and fusing visual and linguistic semantics. We also employ attention mechanisms \cite{zhao2019cam} to focus the linguistic components on the relevant visual features for video captioning. 

\noindent
\textbf{Transformer for Captioning:}
Transformer \cite{vaswani2017attention, khan2022transformers} is increasingly being used for object detection \cite{carion2020end}, classification \cite{han2021transformer}, scene understanding \cite{zhao2021point} and image captioning  \cite{liu2021cptr} tasks. It gained its popularity from the natural language-based models like BERT \cite{devlin2018bert} and roBERTa \cite{liu2019roberta}, a version of BERT \cite{devlin2018bert}, where it demonstrates significant improvement in the performance when using a huge corpus of web text data. These methods are recently adopted for vision-based tasks while using a billion images to give a state-of-the-art performance. \cite{lin2021augmented, lin2022swinbert, seo2022end} are not fair to be compared with as \cite{lin2021augmented} has used an ensemble model together with a transformer decoder and  rest have a pre-training stage for their extremely large transformer-based models using ten to hundred times more computing than ours. Our method outperforms these methods and generates more grammatically correct captions (see Section \ref{ssec:results&comp}).

\subsection{Masking Techniques} 
Previous methods \cite{li2020hero, he2022masked, xie2022simmim} use masking on image frames to generalize the input to the model which improves the performance of the model. Pre-training tasks for language transformers \cite{devlin2018bert, liu2019roberta} require language masking, and similarly, image frames are masked to recover the masked portion of the image. To the best of our knowledge, we are the first to mask spatial features as well as temporal features for an end-to-end video captioning task, which results in an improvement in the performance (see Section \ref{ssec:ablation_block}).
\noindent
\subsection{Visual and Linguistic Alignment and Fusion}
Existing methods \cite{pan2020spatio, wang2019controllable, zhang2019object, zheng2020syntax} generate captions with POS in decoder block without using separate POS blocks. We are the first to explore separate POS blocks in the encoder part to align visual semantics with linguistic POS components in a supervised manner. \cite{kim2021dense} predicts the label of the POS for every word as part of the caption generation using a softmax layer, without any POS blocks. The proposed method aligns the visual and language components at both fine-grained local and global levels by proposing to use of four POS blocks and GLFB.
\section{Method}
This section gives an overview of the proposed method, which is followed by the problem formulation in Section \ref{ssec:num1} and a detailed explanation for the POS blocks in the subsequent sections. \\ 
\noindent
An overview of the proposed network is shown in Figure \ref{fig:2}. The input to the system is a video and the output is a caption that describes the video. Spatial and temporal features are extracted from the input video along with noun features via the nouns anchor block to use as input to the network. We use Deformable DETR \cite{zhu2020deformable, ye2022hierarchical} transformer architecture for the nouns anchor block which takes object features as an input and is supervised by the nouns in the ground-truth caption.
Either spatial or temporal features along with object semantics from the nouns anchor block are given as inputs to the LSTM-based POS blocks using different masking mechanisms.
We have introduced random spatial features masking and chunk-wise temporal features masking as Visual Features Masking (VFM) in Section \ref{ssec:vfm} for an end-to-end video captioning task.
In this work, we proposed 4 POS blocks - Verb, Det + Subject, Aux Verb, and Det + Object. The visual features from the POS blocks are given as input to the GLFB. In addition to this, word embeddings exploiting linguistic information are obtained by supervising POS blocks with respective POS components. 
This helps in achieving the alignment between visual and linguistic features at both the fine-grained local and global levels, unlike previous methods. The fine-grained local and global visual and linguistic features are then fed into the Caption block to generate the final caption.

\subsection{Problem Formulation} \label{ssec:num1}
Given a video-captioning task, we aim to align the visual semantics with the linguistic semantics using the POS blocks. These POS components are inherently present in any grammatically structured caption and these blocks help to exploit fine-grained local visual features of a video to align linguistic and visual semantics. These aligned semantics on the fine-grained local level are then used as input to the GLFB to fuse local and global representations for improved semantic correctness. \\
For supervised training of each POS block, we extract the respective POS component from the ground-truth caption \( \mathcal{C} \). Let \( \mathcal{R} \) be all the reference captions for a given video \( \mathcal{V} \), ground-truth \( \mathcal{G}_{POS} \) for the POS blocks (see Section \ref{ssec:pos_block}) are defined as:
\begin{equation}
\label{eqn:1}
\mathcal{G}_{POS} = \{\{S^{d}, V, A, O^{d}\} \in \mathcal{C}\, |\,  \forall \mathcal{C} \in \mathcal{R} \},
\end{equation}

\noindent
where \( S^{d}, V, A, O^{d} \) are the ground-truth labels for Det + Subject, Verb, Aux Verb and Det + Object blocks respectively. These labels when converted to word embeddings help align the visual semantics with their linguistic counterparts in a supervised setting.
Unlike existing methods that align visual and linguistic semantics on a caption or a nouns-verb level only, we perform a fine-grained local alignment of visual and linguistic semantics for four POS blocks. Our overall objective is as follows:
\begin{equation}
\label{eqn:2}
\mathcal{O}_{OVERALL} = \mathcal{O}_{POS} + \mathcal{O}_{GLFB} + \mathcal{O}_{VFM},
\end{equation}
\noindent
where \( \mathcal{O}_{POS}\), \( \mathcal{O}_{GLFB} \) and \( \mathcal{O}_{VFM} \) are POS, GLFB and VFM objectives, respectively.

\subsection{Visual Features Masking (VFM)}
\label{ssec:vfm}
Random image masking improves the results for image detection and classification tasks in the previous methods \cite{he2022masked, xie2022simmim}. A random patch of the image is masked as an input to the model and the model is forced to recover the masked patch along with the overall image.
In our approach, we randomly mask 30\% of the spatial features and a chunk of the temporal features, approximately 15\%, before the features are input to the POS blocks. The masking percentages are selected through experimentation. When a chunk is masked in the temporal features of a video, it masks the information in the temporal dimension in all the sampled frames. This helps the proposed network model to generalize well across the entire data distribution. It also forces the spatial and temporal features to interact to find the masked information and this improves the overall performance (see Section \ref{ssec:ablation_block}).

\subsection{POS Blocks}
\label{ssec:pos_block}
\begin{figure}[h]
  \centering
   \includegraphics[width=\linewidth]{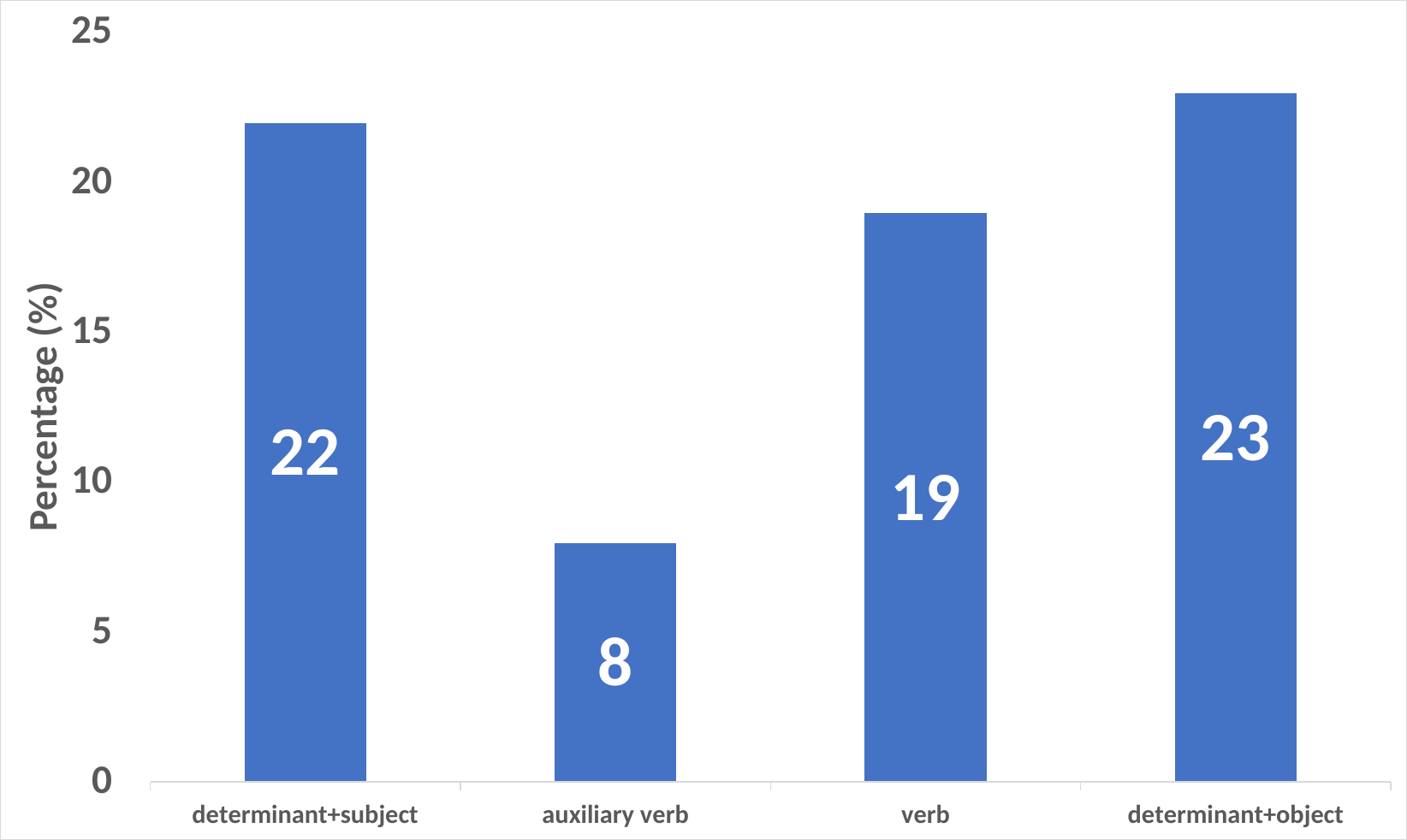}\vspace{-3mm}
   \caption{POS components' representation in the captions from the MSVD and MSRVTT benchmark datasets.}
   \label{fig:4}
\end{figure}

We propose to use four POS blocks in the proposed network, two of these four blocks are a combination of two POS components i.e. 'determinant + subject' and 'determinant + object'. The other two components are 'verb' and 'auxiliary verb'. 
We performed a detailed analysis of the ground-truth captions in benchmark datasets to analyze the presence of the individual parts of speech components in each caption. Figure \ref{fig:4} shows the percentage of the presence of each POS component introduced in our model in the captions for MSVD and MSRVTT benchmark datasets. It is evident from the figure that the four POS blocks used in the proposed method are the largely represented POS components in the captions and also, adding blocks in the network for other less represented POS components affects the performance adversely.  
An extensive ablation (see Section \ref{ssec:ablation_block}) demonstrates the effectiveness of using these four POS blocks within the network to generate grammatically and semantically correct captions. 

\noindent
\textbf{Verb.} This is the first block in the pipeline which takes the temporal features \( \mathnormal{t}_{f} \) as well as the nouns features \( \mathnormal{n}_{f} \) (potential candidates for 'subject' and 'object') via the nouns anchor as input and it aligns the \( \mathnormal{t}_{f} \) and \( \mathnormal{n}_{f} \) with the linguistic verb. These features are concatenated (\( \mathnormal{cat} \)) together and passed through an LSTM layer, in bi-directional mode, to get the visual verb semantics \( \mathnormal{v}_{f} \). Finally, a fully connected (\( \mathnormal{fc} \)) layer is used to predict the linguistic 'verb' embedding \( \mathnormal{v}_{p} \): 
\begin{equation}
\begin{aligned}
\label{eqn:3}
\mathnormal{v}_{f} = \mathnormal{lstm}_{b}(cat(\mathnormal{t}_{f}, \mathcal{A}_{nv}[\mathnormal{n}_{f}])) ,
\end{aligned}
\end{equation}
hence, 
\(\mathnormal{v}_{p} = \mathnormal{fc}(\mathnormal{v}_{f})\). \(\mathcal{A}_{nv}\) is a learnable attention for \( \mathnormal{n}_{f} \). \\
For training of the Verb block, we get 'verb' from the ground-truth caption using pre-trained roBERTa \cite{liu2019roberta} model and use the same model for ground-truth embedding \(\mathnormal{v}_{g}\). The distance with the predicted embedding \(\mathnormal{v}_{p}\) is minimized as part of the model learning. Let \(\mathcal{L}_{v}\) be the Verb block loss:
\begin{equation}
\label{eqn:4}
\mathcal{L}_{v} = \mathnormal{Dist.}( \mathnormal{v}_{p} ,\mathnormal{v}_{g})
\end{equation} 
\noindent
\textbf{Det+Subject.} In this block, we use a combination of two POS components, i.e., 'determinant + subject'. This block uses the spatial visual features \(\mathnormal{s}_{f}\), \( \mathnormal{n}_{f} \) and \( \mathnormal{v}_{f} \) (see Eqn. \ref{eqn:3}) to determine the 'determinant + subject' of the caption. Similar to the Verb block, attention is used to combine the information from the inputs. \( \mathnormal{n}_{f} \) and \( \mathnormal{v}_{f} \) are used as inputs via learnable attentions \(\mathcal{A}_{ns}\) and \(\mathcal{A}_{vs}\) respectively. \(\mathnormal{s}_{f}\) are then concatenated with these attended features and are passed through an LSTM layer similar to the Verb block to get 'determinant + subject' visual features \(\mathnormal{ds}_{f}\). Let \(\mathnormal{ds}_{p}\) be the predicted 'determinant + subject' linguistic embedding after passing \(\mathnormal{ds}_{f}\) through the fully connected layer,
\begin{equation}
\begin{aligned}
\label{eqn:5}
\mathnormal{ds}_{f} = \mathnormal{lstm}_{b}(cat ( \mathnormal{s}_{f}, \mathcal{A}_{vs}[\mathnormal{v}_{f}], \mathcal{A}_{ns}[\mathnormal{n}_{f})]),
\end{aligned}
\end{equation}
hence, 
\(\mathnormal{ds}_{p} = \mathnormal{fc}(\mathnormal{ds}_{f})\).
This approach creates a dynamic template of captions being filled with outputs from the POS blocks as these become available both during training and testing. \\
For labels, we get 'determinant' and 'subject' from the ground-truth caption using the pre-trained roBERTa \cite{liu2019roberta} model. We combine these together with a space in-between and use the same model for ground-truth embedding \(\mathnormal{ds}_{g}\). The distance with the predicted embedding \(\mathnormal{ds}_{p}\) is minimized as a part of the model learning. Let \(\mathcal{L}_{ds}\) be the Det+Subject block loss:
\begin{equation}
\label{eqn:6}
\mathcal{L}_{ds} = \mathnormal{Dist.}( \mathnormal{ds}_{p} ,\mathnormal{ds}_{g})
\end{equation}
\noindent
\textbf{Aux Verb.} Aux verb block uses \( \mathnormal{t}_{f} \), noun features \( \mathnormal{n}_{f} \), \( \mathnormal{v}_{f} \) (see Eqn. \ref{eqn:3}) and \(\mathnormal{ds}_{f}\) (see Eqn. \ref{eqn:5}) to predict 'auxiliary verb' for the caption. It also gives an insight into the singular or plural nature of the 'subject', which is used as an input to the GLFB. 
\( \mathnormal{n}_{f} \),  \( \mathnormal{v}_{f} \) and \(\mathnormal{ds}_{f}\) are used as the inputs via learnable attentions \(\mathcal{A}_{na}\), \(\mathcal{A}_{va}\) and \(\mathcal{A}_{dsa}\) respectively. \( \mathnormal{t}_{f} \) is then concatenated with these attended features and is passed through an LSTM layer like in previous blocks to get 'auxiliary verb' visual semantics \(\mathnormal{a}_{f}\). \(\mathnormal{a}_{p}\) is obtained after passing \(\mathnormal{a}_{f}\) through a fully connected layer:
\begin{equation}
\begin{aligned}
\label{eqn:7}
\mathnormal{a}_{f} = \mathnormal{lstm}_{b}(cat ( \mathnormal{t}_{f},  \mathcal{A}_{va}[\mathnormal{v}_{f}], \mathcal{A}_{na}[\mathnormal{n}_{f}], \mathcal{A}_{dsa}[\mathnormal{ds}_{f}])),
\end{aligned}
\end{equation}

\noindent
hence, 
\(\mathnormal{a}_{p} = \mathnormal{fc}(\mathnormal{a}_{f})\).
For ground-truth labels, we extract the 'auxiliary verb' from the ground-truth caption and also, its embedding \(\mathnormal{a}_{g}\) using pre-trained roBERTa \cite{liu2019roberta} model. Then, the distance with \(\mathnormal{a}_{p}\) is minimized as part of the model training. Let \(\mathcal{L}_{a}\) be the Aux Verb block loss:
\begin{equation}
\label{eqn:8}
\mathcal{L}_{a} = \mathnormal{Dist.}( \mathnormal{a}_{p} ,\mathnormal{a}_{g})
\end{equation} 

\noindent
\textbf{Det+Object.} Last, in the POS blocks, the Det+Object block is a combination of the 'determinant' and 'object'. It exploits linguistically aligned visual features from the previous POS blocks along with \( \mathnormal{s}_{f} \). Like previous POS blocks, learnable attentions \(\mathcal{A}_{no}\), \(\mathcal{A}_{vo}\), \(\mathcal{A}_{dso}\) and \(\mathcal{A}_{ao}\) are used to exploit information in \( \mathnormal{n}_{f} \), \( \mathnormal{v}_{f} \), \(\mathnormal{ds}_{f}\) and \(\mathnormal{a}_{f}\) respectively. In similar manner, \(\mathnormal{do}_{p}\) is obtained from \(\mathnormal{do}_{f}\) by:
\begin{equation}
\begin{aligned}
\label{eqn:9}
\mathnormal{do}_{f} = \mathnormal{lstm}_{b}(cat ( \mathnormal{s}_{f}, \mathcal{A}_{vo}[\mathnormal{v}_{f}], \mathcal{A}_{no}[\mathnormal{n}_{f}],\\ \mathcal{A}_{dso}[\mathnormal{ds}_{f}], \mathcal{A}_{ao}[\mathnormal{a}_{f}])),
\end{aligned}
\end{equation}
\noindent
hence, 
\(\mathnormal{do}_{p} = \mathnormal{fc}(\mathnormal{do}_{f})\).
For training, we get the 'determinant' and 'object' from the ground-truth caption using pre-trained roBERTa \cite{liu2019roberta} model, combine these together with a space in-between and then, use the same model for ground-truth embedding \(\mathnormal{do}_{g}\). The distance with the predicted embedding \(\mathnormal{do}_{p}\) is minimized as part of the model learning. Let \(\mathcal{L}_{do}\) be the Det+Object block loss: 
\begin{equation}
\label{eqn:10}
\mathcal{L}_{do} = \mathnormal{Dist.}( \mathnormal{do}_{p} ,\mathnormal{do}_{g})
\end{equation} 

\subsection{Global Local Fusion Block (GLFB)}
\label{ssec:glcr_block}
The proposed GLFB is used to achieve fusion between the global and local features, by using the fine-grained local  and linguistically aligned visual features from the POS blocks and the visual-spatial features.
Visual features \(\mathnormal{g}_{f}\) are the output of an LSTM layer, given a set of inputs using the attention mechanism. To predict a linguistic GLFB representation \(\mathnormal{g}_{p}\), \(\mathnormal{g}_{f}\) is passed through a fully connected layer:
\begin{equation}
\begin{aligned}
\label{eqn:11}
\mathnormal{g}_{f} = \mathnormal{lstm}_{b}(cat ( \mathnormal{s}_{f},  \mathcal{A}_{vg}[\mathnormal{v}_{f}], \mathcal{A}_{ng}[\mathnormal{n}_{f}],\\ \mathcal{A}_{dsg}[\mathnormal{ds}_{f}], \mathcal{A}_{ag}[\mathnormal{a}_{f}],
\mathcal{A}_{dog}[\mathnormal{do}_{f}])),
\end{aligned}
\end{equation}

\noindent
hence, 
\(\mathnormal{g}_{p} = \mathnormal{fc}(\mathnormal{g}_{f})\).
For training of this block, the whole ground-truth caption is first converted to lower-case letters and then, passed through roBERTa \cite{liu2019roberta} to get ground-truth embedding \(\mathnormal{g}_{g}\). Then, the distance with the predicted embedding \(\mathnormal{g}_{p}\) is minimized as part of the supervised learning. Let \(\mathcal{L}_{g}\) be the GLFB loss, then:
\begin{equation}
\label{eqn:12}
\mathcal{L}_{g} = \mathnormal{Dist.}( \mathnormal{g}_{p} ,\mathnormal{g}_{g})
\end{equation} 

\subsection{Attention Mechanism}
Our approach employs an attention mechanism in all blocks to exploit the inputs as can be seen in previous sections. Here, we generically define the attention \(\alpha_{g}\):

\begin{equation}
\begin{aligned}
\label{eqn:15}
\alpha_{g} = \mathnormal{w}_{g}\,\mathnormal{ tanh}(\mathcal{W}_{g}\,\mathnormal{X} + \mathcal{U}_{g}\,\mathnormal{Y} + \mathnormal{b}_{g}), 
\end{aligned}
\end{equation}

\noindent
where \(\mathnormal{w}_{g}\), \(\mathcal{W}_{g}\), \(\mathcal{U}_{g}\) and \(\mathnormal{b}_{g}\) are the training parameters. \(\mathnormal{X}\) is either spatial or temporal features, depending on whatever is the input to each block, in all blocks, except the Caption block (see Section \ref{ssec:OTO}) where it is the previous word in the generated caption. \(\mathnormal{Y}\) is the visual semantics generated in the POS blocks and GLFB.

\subsection{Overall Training Objective}
\label{ssec:OTO}
The overall training objective adds the POS blocks and GLFB together along with the Caption block. The Caption block uses an LSTM layer, followed by a fully connected layer, with inputs of both visual and linguistic semantics from the POS blocks and GLFB  to generate a caption, word by word, as an output. Overall learning loss \(\mathcal{L}_{all}\) is defined as:
\begin{equation}
\begin{aligned}
\label{eqn:13}
\mathcal{L}_{all} = \mathcal{L}_{c}+\mathcal{L}_{v}+\mathcal{L}_{ds}+\mathcal{L}_{a}+\mathcal{L}_{do}+\mathcal{L}_{g},
\end{aligned}
\end{equation}
where \(\mathcal{L}_{c}\) is the cross entropy loss for the Caption block:
\begin{equation}
\label{eqn:14}
\mathcal{L}_{c} = -\displaystyle\sum_{n=1} ^{N} E(\mathnormal{w}_{n}) \mathnormal{log} \mathnormal{P}_{n}
\end{equation}

\noindent
\(E(\mathnormal{w}_{n})\) is the one-hot encoding of the word \(\mathnormal{w}_{n}\) and \(\mathnormal{P}_{n}\), the output of the fully connected layer, is the probability distribution over the word vocabulary. It is worth mentioning that loss for the VFM objective is an integral part of \(\mathcal{L}_{all}\) as the model has to predict the right embedding with masked input. Also, we have done some experiments by tuning the loss weights but the effect on the overall performance was found to be minimal.\\
\noindent
The objective of the Caption block is to generate the final video caption which is both grammatically and semantically correct. 

\section{Experimental Results and Evaluations}
\noindent
\textbf{Datasets.} Similar to the existing video captioning approaches \cite{zhang2019object, pei2019memory, wang2019controllable, aafaq2019spatio, pan2020spatio, zheng2020syntax, ryu2021semantic, ye2022hierarchical}, we evaluate our model on MSRVTT \cite{xu2016msr} and MSVD \cite{chen2011collecting} benchmark datasets. These datasets have a total of 10,000 and 1,970 videos respectively. We use the same train, validation, and test split as the existing methods. \\
\textbf{Metrics.} For evaluation with the existing methods, we use the following set of video captioning metrics: BLEU@4 (B@4)\cite{papineni2002bleu}, METEOR (M)\cite{banerjee2005meteor}, ROUGE-L (R)\cite{lin2004rouge} and CIDEr (C)\cite{vedantam2015cider}. CIDEr \cite{vedantam2015cider} is studied to be robust in the condition where the semantic meaning of the caption remains intact \cite{caglayan2020curious}. Also, we are the first to use GPT-2 \cite{radford2019language} pre-trained model for measuring the Grammatical correctness Score (GS) of the captions generated by our model in comparison to state-of-the-art which we implement for the GS metric, demonstrating improved performance (see Section \ref{ssec:ablation_block}). \\
\textbf{Implementation.} For text, we have used \emph{spaCy}\footnote{https://spacy.io/} with roBERTa \cite{liu2019roberta}, a version of BERT \cite{devlin2018bert}, to extract POS components along with nouns from the ground-truth captions. We again use pre-trained roBERTa \cite{liu2019roberta} for text embedding which gives an embedding of size 768 for 'determinant + subject', 'verb', 'auxiliary verb', 'determinant + object', and the whole caption. Following the existing methods, we use InceptionResNetV2 \cite{szegedy2017inception} to extract the spatial features and C3D \cite{hara2018can} to extract the temporal features. These features are projected to 512 sizes before being input into the network. We train for epochs 25 and use a learning rate of 0.00015, batch size 16, ADAM optimizer \cite{kingma2014adam}, 16 samples per video as well as a hidden state size of 512 for the Caption block. Our model has 76M parameters and 0.045s inference time. Apart from that, we use Yolov7 \cite{wang2022yolov7} for extracting object features for the noun anchor. The whole implementation is performed using one NVIDIA GeForce RTX 3090 and PyTorch.

\begin{table*}[h!]
\centering
\resizebox{\textwidth}{!}{%
\begin{tabular}{*{1}c|*{1}c|*{5}c|*{5}c}
\hline
Method & Language Components & \multicolumn{5}{c|}{MSVD}                                                                                                     & \multicolumn{5}{c}{MSRVTT}                                                                                                  \\ \cline{3-12} 
                        &                                      &\multicolumn{1}{c}{B@4 \(\uparrow\)}    & \multicolumn{1}{c}{M \(\uparrow\)}    & \multicolumn{1}{c}{R  \(\uparrow\)}    & \multicolumn{1}{c}{C  \(\uparrow\)} & GS  \(\downarrow\)                         & \multicolumn{1}{c}{B@4 \(\uparrow\)}    & \multicolumn{1}{c}{M \(\uparrow\)}    & \multicolumn{1}{c}{R  \(\uparrow\)}    & \multicolumn{1}{c}{C  \(\uparrow\)} & GS  \(\downarrow\)                    \\ \hline
OA-BTG \cite{zhang2019object}                 & Nouns (Physical objects)             & \multicolumn{1}{c}{56.9}          & \multicolumn{1}{c}{36.2}          & \multicolumn{1}{c}{-} & \multicolumn{1}{c}{90.6}            & 852.6           & \multicolumn{1}{c}{41.4}          & \multicolumn{1}{c}{28.2}          & \multicolumn{1}{c}{-}         & \multicolumn{1}{c}{46.9} & 337.5          \\ \hline 
MARN \cite{pei2019memory}                   & Word by word (Global)                & \multicolumn{1}{c}{48.6}          & \multicolumn{1}{c}{35.1}          & \multicolumn{1}{c}{71.9}          & \multicolumn{1}{c}{92.2}   & 822.2        & \multicolumn{1}{c}{40.4}          & \multicolumn{1}{c}{28.1}          & \multicolumn{1}{c}{60.7}          & \multicolumn{1}{c}{47.1} & 328.1         \\ \hline 
POS-CG \cite{wang2019controllable}                 & Subject-verb-object                  & \multicolumn{1}{c}{52.5}          & \multicolumn{1}{c}{34.1}          & \multicolumn{1}{c}{71.3}   & \multicolumn{1}{c}{88.7}       & 801.5           & \multicolumn{1}{c}{42.0}          & \multicolumn{1}{c}{28.2}          & \multicolumn{1}{c}{61.6} & \multicolumn{1}{c}{48.7} & 306.4          \\ \hline 
GRU-EVE \cite{aafaq2019spatio}                 & Nouns-verb                           & \multicolumn{1}{c}{47.9}          & \multicolumn{1}{c}{35.0}          & \multicolumn{1}{c}{71.5}          & \multicolumn{1}{c}{78.1}  & -           & \multicolumn{1}{c}{38.3}          & \multicolumn{1}{c}{28.4}          & \multicolumn{1}{c}{60.7}          & \multicolumn{1}{c}{48.1} & -          \\ \hline 
STG-KD \cite{pan2020spatio}                  & Nouns-verb                           & \multicolumn{1}{c}{52.2}          & \multicolumn{1}{c}{36.9}          & \multicolumn{1}{c}{73.9} & \multicolumn{1}{c}{93.0} & 802.9           & \multicolumn{1}{c}{40.5}          & \multicolumn{1}{c}{28.3}          & \multicolumn{1}{c}{60.9}          & \multicolumn{1}{c}{47.1} & 300.3          \\ \hline 
SAAT \cite{zheng2020syntax}                   & Subject-verb-object                  & \multicolumn{1}{c}{46.5}          & \multicolumn{1}{c}{33.5}          & \multicolumn{1}{c}{69.4}          & \multicolumn{1}{c}{81.0} & 806.1           & \multicolumn{1}{c}{40.5}          & \multicolumn{1}{c}{28.2}          & \multicolumn{1}{c}{60.9}          & \multicolumn{1}{c}{49.1} & 282.6          \\ \hline 
SGN \cite{ryu2021semantic}                    & Word by word (Global)                & \multicolumn{1}{c}{52.8}          & \multicolumn{1}{c}{35.5}          & \multicolumn{1}{c}{72.9}          & \multicolumn{1}{c}{94.3} & 830.3           & \multicolumn{1}{c}{40.8}          & \multicolumn{1}{c}{28.3}          & \multicolumn{1}{c}{60.8}          & \multicolumn{1}{c}{49.5} & 316.4         \\ \hline 
GL-RG \cite{yan2022gl}                    & Word by word (Global)                & \multicolumn{1}{c}{55.5}          & \multicolumn{1}{c}{37.8}          & \multicolumn{1}{c}{74.7}          & \multicolumn{1}{c}{94.3} & 792.5           & \multicolumn{1}{c}{45.5}          & \multicolumn{1}{c}{30.1}          & \multicolumn{1}{c}{62.6}          & \multicolumn{1}{c}{51.2} & 247.1         \\ \hline 
HMN \cite{ye2022hierarchical}                    & Nouns-verb                & \multicolumn{1}{c}{59.2}          & \multicolumn{1}{c}{37.7}          & \multicolumn{1}{c}{75.1}          & \multicolumn{1}{c}{104.0} & 728.7           & \multicolumn{1}{c}{43.5}          & \multicolumn{1}{c}{29.0}          & \multicolumn{1}{c}{62.7}          & \multicolumn{1}{c}{51.5} & 214.7          \\ \hline 
SEM-POS (Ours)          & Subject-aux verb-verb-object              & \multicolumn{1}{c}{\textbf{60.1}} & \multicolumn{1}{c}{\textbf{38.5}} & \multicolumn{1}{c}{\textbf{76.0}} & \multicolumn{1}{c}{\textbf{108.3}} & \textbf{607.1} & \multicolumn{1}{c}{\textbf{45.2}} & \multicolumn{1}{c}{\textbf{30.7}} & \multicolumn{1}{c}{\textbf{64.1}} & \multicolumn{1}{c}{\textbf{53.1}} & \textbf{192.6} \\ \hline
\end{tabular}%
}
\parbox[t]{\textwidth}{\caption{Comparison against state-of-the-art methods. The best results on each metric are in bold on MSVD and MSRVTT benchmarks.} 
\label{table:1}\vspace{-6mm}}
\end{table*}
\subsection{Results and Comparison with Existing Methods}
\label{ssec:results&comp}
We do a quantitative evaluation of the benchmark datasets for several state-of-the-art methods and also, demonstrate qualitative results of the proposed method.

\noindent
\textbf{Quantitative Evaluation:}
Table \ref{table:1} gives a quantitative comparative performance of the proposed model against the state-of-the-art methods which do not employ large-scale transformer pre-training. 
The proposed SEM-POS network outperforms all these methods on both MSVD \cite{chen2011collecting} and MSRVTT \cite{xu2016msr} benchmarks. Improvement is around 3\% and 4\% on CIDEr for MSRVTT and MSVD benchmarks respectively. In terms of GS metric, our method gives an improvement of around 10\% and 16\% for MSRVTT and MSVD respectively.\\
This is due to the fact that our method explores fine-grained local features in the four POS blocks before generating a global representation, giving it a better generalization ability compared to the other approaches. Visual features aligned with linguistics in POS blocks are learned across the entire data distribution for the video captioning task, which contributes to this improved performance. POS blocks help to align the linguistic and visual semantics at the fine-grained local level and then the GLFB visibly fuses it with global representation. Both these contributions of POS blocks and GLFB help generate grammatically and semantically correct sentences. Our method also outperforms these methods \cite{lin2021augmented, lin2022swinbert, seo2022end} on GS metric which have a large-scale transformer-based pre-training stage and have 655.2, 643.7, and 612.8 (MSVD) and 224.7, 212.9 and 201.6 (MSRVTT) scores respectively.   

\begin{figure}[t]
  \centering
   \includegraphics[width=\linewidth]{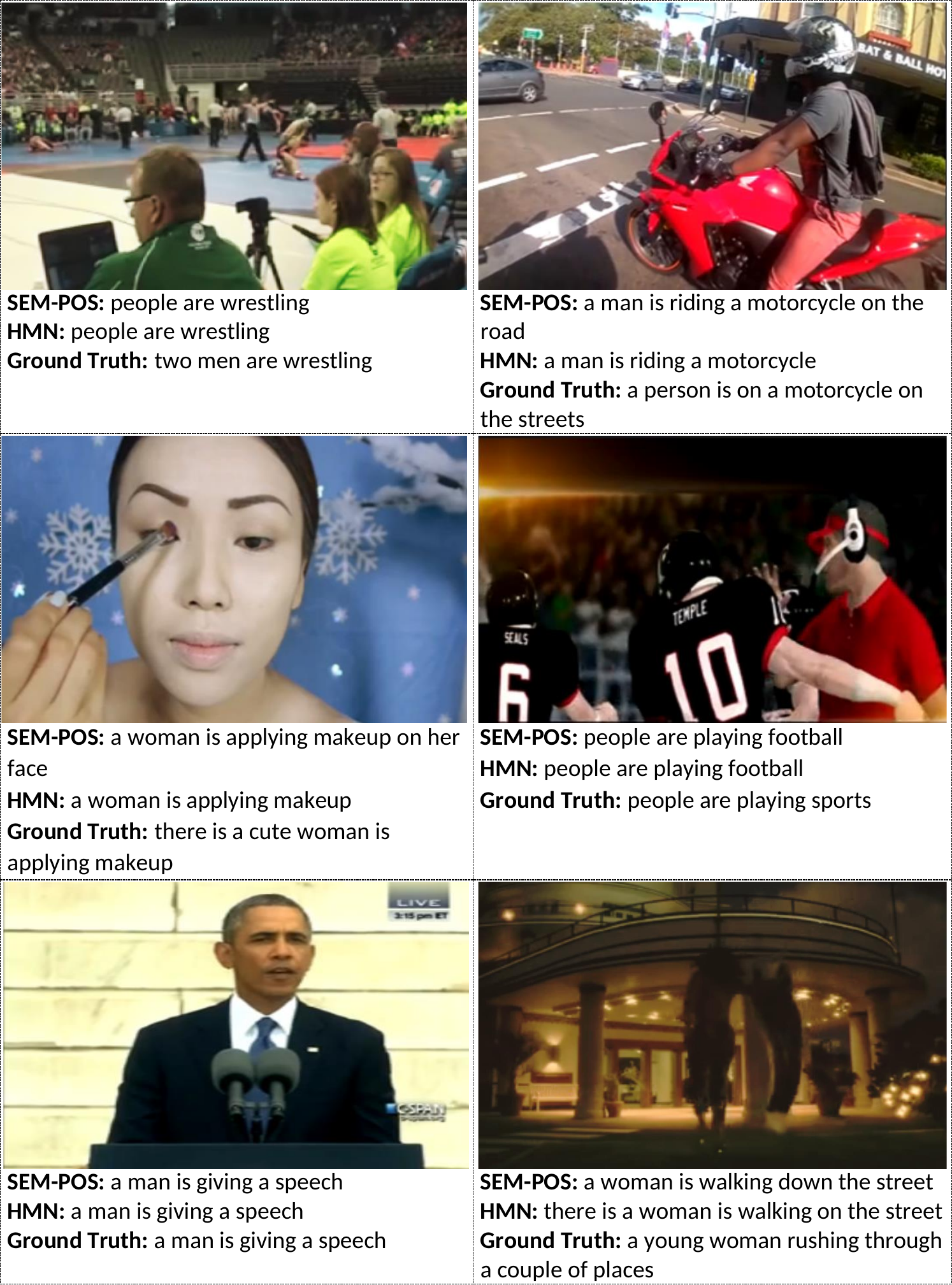}
   \caption{Qualitative Results- SEM-POS predictions show fine-grained visual details and global-local features' fusion.}
   \label{fig:3}
\end{figure}

\noindent
\textbf{Qualitative Results:}
In Figure \ref{fig:3}, we show qualitative results for our approach. It can be seen that generated captions adequately represent not only the ground-truth captions but also the fine-grained visual features. For example, in the images in the first two rows, on the right-hand side, the proposed caption is better than ground truth and provides more specific information about 'road' and 'football' instead of 'streets' and 'sports' respectively, which is derived from the fine-grained local features in the POS blocks. \\ We also compare the qualitative results of our method with the state-of-the-art method HMN \cite{ye2022hierarchical}. Row 1 right, row 2 left and row 3 right show that our method generates captions that are grammatically and  semantically more correct than HMN \cite{ye2022hierarchical}. 
The rest of the results also show that our method generates captions as well as state-of-the-art. 

\begin{table*}[h!]
\centering
\resizebox{\textwidth}{!}{%
\begin{tabular}{*{1}c|*{5}c|*{5}c}
\hline
Block         & \multicolumn{5}{c|}{MSVD}                                                                                      & \multicolumn{5}{c}{MSRVTT}                                                                                   \\ \cline{2-11}  
                               &\multicolumn{1}{c}{B@4 \(\uparrow\)}    & \multicolumn{1}{c}{M \(\uparrow\)}    & \multicolumn{1}{c}{R  \(\uparrow\)}    & \multicolumn{1}{c}{C  \(\uparrow\)} & GS  \(\downarrow\)                         & \multicolumn{1}{c}{B@4 \(\uparrow\)}    & \multicolumn{1}{c}{M \(\uparrow\)}    & \multicolumn{1}{c}{R  \(\uparrow\)}    & \multicolumn{1}{c}{C  \(\uparrow\)} & GS  \(\downarrow\)                    \\ \hline
w/o Det + Subject                & \multicolumn{1}{c}{56.4}&\multicolumn{1}{c}{36.6} & \multicolumn{1}{c}{72.0} & \multicolumn{1}{c}{100.1} & 695.1                      & \multicolumn{1}{c}{41.2} & \multicolumn{1}{c}{28.0} & \multicolumn{1}{c}{60.0} & \multicolumn{1}{c}{49.8} & 244.9                     \\ \hline
w/o Verb                       & \multicolumn{1}{c}{55.9} & \multicolumn{1}{c}{34.9} & \multicolumn{1}{c}{70.8} & \multicolumn{1}{c}{98.1} & 733.1                      & \multicolumn{1}{c}{40.8} & \multicolumn{1}{c}{27.6} & \multicolumn{1}{c}{58.3} & \multicolumn{1}{c}{49.1} & 267.3                     \\ \hline 
w/o Aux Verb                   & \multicolumn{1}{c}{57.7} & \multicolumn{1}{c}{37.1} & \multicolumn{1}{c}{73.9} & \multicolumn{1}{c}{101.2} & 688.3                     & \multicolumn{1}{c}{42.5} & \multicolumn{1}{c}{28.2} & \multicolumn{1}{c}{60.9} & \multicolumn{1}{c}{50.6} &  244.2                   \\ \hline 
w/o Det + Object                 & \multicolumn{1}{c}{57.8} & \multicolumn{1}{c}{37.1} & \multicolumn{1}{c}{73.9} & \multicolumn{1}{c}{102.3} & 642.4                     & \multicolumn{1}{c}{41.6} & \multicolumn{1}{c}{28.9} & \multicolumn{1}{c}{62.1} & \multicolumn{1}{c}{51.7} & 221.1                     \\ \hline
w/o GLFB & \multicolumn{1}{c}{56.7} & \multicolumn{1}{c}{36.0} & \multicolumn{1}{c}{73.1} & \multicolumn{1}{c}{101.2}  & 638.1 & \multicolumn{1}{c}{41.9} & \multicolumn{1}{c}{28.2} & \multicolumn{1}{c}{61.9} & \multicolumn{1}{c}{50.1} & 246.3 \\ \hline

w/o VFM  & \multicolumn{1}{c}{59.6} & \multicolumn{1}{c}{38.1} & \multicolumn{1}{c}{75.3} & \multicolumn{1}{c}{106.5} & 628.7 & \multicolumn{1}{c}{44.6} & \multicolumn{1}{c}{29.6} & \multicolumn{1}{c}{63.4} & \multicolumn{1}{c}{52.5} & 224.1 \\ \hline
w Adverb                & \multicolumn{1}{c}{56.5} & \multicolumn{1}{c}{36.8} & \multicolumn{1}{c}{71.2} & \multicolumn{1}{c}{99.4} & 710.1                      & \multicolumn{1}{c}{40.7} & \multicolumn{1}{c}{28.1} & \multicolumn{1}{c}{59.5} & \multicolumn{1}{c}{49.6} & 264.9                     \\ \hline
w Adjective                       & \multicolumn{1}{c}{56.2} & \multicolumn{1}{c}{34.5} & \multicolumn{1}{c}{70.8} & \multicolumn{1}{c}{98.1} & 762.3                      & \multicolumn{1}{c}{40.8} & \multicolumn{1}{c}{27.8} & \multicolumn{1}{c}{57.5} & \multicolumn{1}{c}{49.3} & 273.6                     \\ \hline
w Conjunction                   & \multicolumn{1}{c}{56.4} & \multicolumn{1}{c}{35.3} & \multicolumn{1}{c}{72.9} & \multicolumn{1}{c}{97.5} & 785.4                     & \multicolumn{1}{c}{40.4} & \multicolumn{1}{c}{27.7} & \multicolumn{1}{c}{59.1} & \multicolumn{1}{c}{49.0} &  275.7                    \\ \hline
w Subject only                 & \multicolumn{1}{c}{58.4} & \multicolumn{1}{c}{37.9} & \multicolumn{1}{c}{74.7} & \multicolumn{1}{c}{103.3} & 640.1                     & \multicolumn{1}{c}{43.8} & \multicolumn{1}{c}{28.9} & \multicolumn{1}{c}{62.6} & \multicolumn{1}{c}{51.5} & 206.1                     \\ \hline
w Object only               & \multicolumn{1}{c}{58.9} & \multicolumn{1}{c}{37.8} & \multicolumn{1}{c}{74.9} & \multicolumn{1}{c}{103.7}  &   632.5                 & \multicolumn{1}{c}{44.0} & \multicolumn{1}{c}{29.1} & \multicolumn{1}{c}{61.6} & \multicolumn{1}{c}{51.7} &  209.5                   \\ \hline
w All & \multicolumn{1}{c}{57.9} & \multicolumn{1}{c}{35.9} & \multicolumn{1}{c}{73.1} & \multicolumn{1}{c}{100.5}  & 657.3 & \multicolumn{1}{c}{41.4} & \multicolumn{1}{c}{27.5} & \multicolumn{1}{c}{61.2} & \multicolumn{1}{c}{48.4} & 293.6 \\ \hline
SEM-POS (Ours)  & \multicolumn{1}{c}{\textbf{60.1}} & \multicolumn{1}{c}{\textbf{38.5}} & \multicolumn{1}{c}{\textbf{76.0}} & \multicolumn{1}{c}{\textbf{108.3}} & \textbf{607.1} & \multicolumn{1}{c}{\textbf{45.2}} & \multicolumn{1}{c}{\textbf{30.7}} & \multicolumn{1}{c}{\textbf{64.1}} & \multicolumn{1}{c}{\textbf{53.1}} & \textbf{192.6} \\ \hline
\end{tabular}%
}
\parbox[t]{\textwidth}{\caption{Ablation results for the effectiveness of POS, GLFB, and VFM on MSVD and MSRVTT benchmarks.}
\label{table:3}\vspace{-6mm}}
\end{table*}



%
\subsection{Ablation Results}
\label{ssec:ablation_block}
\noindent
\textbf{Effectiveness of POS blocks and GLFB.} 
This section supports our first two contributions of using POS blocks and GLFB in the network and the results are shown in Table \ref{table:3}. Ablation is performed on the proposed method by removing the POS blocks one by one from the proposed network and evaluating the performance. The first four rows in the table demonstrate these results. As it is evident from the table the Verb block contributes significantly to the performance of the network. Det+Subject and Det+Object blocks contribute equally to the MSRVTT \cite{xu2016msr} benchmark but Det+Object shows only a  minor contribution for the MSVD \cite{chen2011collecting} benchmark dataset as it has a significant number of ground-truth captions without an object.

\noindent
'w/o GLFB' shows results without the GLFB (see Section \ref{ssec:glcr_block}). 
It demonstrates the importance of combining fine-grained local and global features for video captioning.
Three rows 'w Adverb', 'w Adjective', and 'w Conjunction' show that the performance is affected adversely as these POS components are not widely represented in the captions. Rows 'w Subject only' and 'w Object only' are Subject and Object blocks without Determinant and these show a decrease in performance. \\
Row 'w All' show the results for adding Adverb, Adjective, and Conjunction blocks to our model. It is evident that adding blocks that are not represented enough in the POS components, decreases the performance significantly.

\noindent
\textbf{Effectiveness of VFM:} 
We also evaluate the performance without Visual Feature Masking (VFM) which is introduced in Section \ref{ssec:vfm}. As seen from 'w/o VFM' in Table \ref{table:3}, masking spatial and temporal features increases the performance to a considerable extent. We use different techniques of masking for spatial and temporal features. For spatial features, we find random masking across the whole frame features to be more effective than masking particular patches in a frame. We have introduced chunk-wise masking for temporal features as random masking is counterproductive in this case. In chunk-wise masking, we mask a chunk of values that are adjacent to each other in the temporal dimension and spatially the same for each frame.

\subsection{Grammatical Score Evaluation}
In video captioning, in addition to semantic correctness, it is important to check whether the generated captions are grammatically correct.
Existing methods only evaluate the semantic correctness of the generated captions, however, in this section, we evaluate the Grammatical Score (GS) of the captions generated from the proposed method against state-of-the-art methods.
\cite{WinNT} performed a comprehensive study to use GPT-2 \cite{radford2019language} for measuring the grammatical correctness of English sentences. We employ the same technique to measure the average GS of generated captions on both MSVD \cite{chen2011collecting} and MSRVTT \cite{xu2016msr} datasets as part of our ablation study in Table \ref{table:3} and we compare the GS against state-of-the-art methods in Table \ref{table:1}. It is based on the perplexity measurement for each caption and the lower score means that the caption has fewer grammatical errors. The results demonstrate that the proposed method with POS blocks and GLFB blocks gives the best performance and generated more descriptive and grammatically correct captions compared to the existing methods achieving the new state-of-the-art.   
\subsection{Limitations}
The performance of each POS block is limited to the number of samples present in the datasets. For e.g., due to a lack of 'auxiliary verb' samples in the dataset, this does not contribute significantly to the final performance. Similarly, other POS, like adverbs and adjectives, are far less represented in the captions, and adding these blocks, affects the performance adversely. Also the proposed method does not handle multiple subjects/objects like existing methods. We aim to address these limitations in our future work.
\section{Conclusion}
This paper introduces a novel global-local feature fusion network that fuses local and global, language and visual features for video captioning. Four POS and a GLFB block is introduced within the network to create a coherent representation to generate grammatically and semantically correct captions, which is proved by the performance of the proposed method on MSVD \cite{chen2011collecting} and MSRVTT \cite{xu2016msr} benchmarks for five different metrics against state-of-the-art methods.
Application of Visual Features Masking (VFM) within the novel network helps our model to not only achieve better performance but also generalize well across the data distribution, as seen in the ablation results.
Qualitative results show the ability of our model to learn fine-grained details in the video and we believe that the insights from this work will further open up research in video-language alignment and understanding not only for semantic correctness but also for grammatical correctness.
\section{Acknowledgement}
This research was partly supported by the British Broadcasting Corporation Research and Development (BBC R\&D), Engineering and Physical Sciences Research Council (EPSRC) Grant EP/V038087/1 “BBC Prosperity Partnership: Future Personalised Object-Based Media Experiences Delivered at Scale Anywhere”.




{\small
\bibliographystyle{ieee_fullname}
\bibliography{egbib}
}

\end{document}